\pdfoutput=1

\documentclass[11pt]{article}

\usepackage{EMNLP2023}

\usepackage{times}
\usepackage{latexsym}

\usepackage[T1]{fontenc}

\usepackage[utf8]{inputenc}

\usepackage{microtype}

\usepackage{inconsolata}

\usepackage{gb4e}
\noautomath
\usepackage{multirow}
\usepackage{multicol}
\usepackage{tikz}
\usepackage{adjustbox}
\usepackage{booktabs}
\usepackage{amsfonts}
\usepackage{cuted}
\usepackage{mathtools,amssymb}
\usepackage{forest}

\title{Taxonomic Loss for Morphological Glossing of Low-Resource Languages}

\author{Michael Ginn \and Alexis Palmer \\ University of Colorado \\
  \texttt{michael.ginn@colorado.edu} \and \texttt{alexis.palmer@colorado.edu} \\}

\begin{document}
\maketitle
\begin{abstract}
Morpheme glossing is a critical task in automated language documentation and can benefit other downstream applications greatly. While state-of-the-art glossing systems perform very well for languages with large amounts of existing data, it is more difficult to create useful models for low-resource languages. In this paper, we propose the use of a taxonomic loss function that exploits morphological information to make morphological glossing more performant when data is scarce. We find that while the use of this loss function does not outperform a standard loss function with regards to single-label prediction accuracy, it produces better predictions when considering the top-n predicted labels. We suggest this property makes the taxonomic loss function useful in a human-in-the-loop annotation setting.
\end{abstract}

\section{Introduction}
With over half of the world's languages at risk of extinction, there is motivation to develop language technology and models that are effective when resources are scarce \citep{seifart_language_2018}. Particularly, developing automated systems for language documentation can greatly aid in preservation efforts \citep{zhao_automatic_2020, mcmillan-major_automating_2020}. However, the languages most in need of documentation rarely have a sufficient amount of annotated text to train effective models. 

While acquiring large amounts of additional data is often infeasible, many languages have dedicated experts who have conducted linguistic analyses and created resources describing the morphology, grammar, phonology, and other aspects of the language. This work focuses on morphological resources that describe classes of morphemes that occur in similar contexts in the form of a \textbf{morphological taxonomy}.

We utilize this information to define a \textbf{taxonomic loss function}, where the model is rewarded for predicting more similar glosses and discouraged from predicting unrelated ones. We find that the use of this loss function can improve performance under certain conditions, and is particularly well-suited to a scenario where the model suggests multiple possible glosses for a human annotator to choose from.

The code is available on GitHub\footnote{\url{https://github.com/michaelpginn/taxo-morph/tree/emnlp-2023}}.

\section{Background}
\subsection{Task}
\textbf{Morpheme glossing} refers to the labeling of each morpheme in a string of text with the function or translation of the morpheme. Morpheme glossing is a key aspect in the creation of the interlinear glossed text (IGT) format for language documentation.

In IGT, morphemes which perform a grammatical function are glossed with a tag for their function; for instance, the English plural morpheme \textit{-s} would be glossed as \textsc{Pl}. Stem morphemes, which encode semantic information, may be glossed with their part of speech or with their translation; for instance, the Latin stem \textit{vin-} would be glossed as a noun: \textsc{N}, or with its translation: \textsc{wine}. Altogether, a complete gloss line of IGT might look like \ref{igt}.

\begin{small}
  \begin{exe}
    \ex 
    \gll ní-s-nith anúnas \\
    Neg.3sf.\textit{eats} {from above}\\
    \citep{lewis_developing_2010}
    \label{igt}
  \end{exe}
  \end{small}

For simplicity, our model learns to gloss using only part-of-speech tags for stems.

\subsection{Related Work}
A number of approaches have been used to create systems for automated glossing. These include rule-based methods \citep{bender2014learning}, active learning \citep{palmer_computational_2010, palmer2009evaluating}, conditional random fields \citep{moeller_automatic_2018, mcmillan-major_automating_2020}, and neural models \citep{moeller_automatic_2018, zhao_automatic_2020}. However, to our knowledge no one has attempted to integrate explicit morphological information into a neural model.

The approach used in this research is inspired by \citet{wu_deep_2019} and \citet{nourani-vatani_structured_2015}, which use structured loss functions to improve classification for plant identification and seafloor image recognition, respectively. However, to our knowledge, such a loss function has never been used in the context of morphology or language.

\section{Data \& Methodology}
\subsection{Data}
The experiments in this research use an IGT corpus of Uspanteko data (Mayan, Guatemala) originally from OKMA documentation project \citep{pixabaj2007text} and adapted by \citet{palmer2009evaluating}. The corpus consists of 9,774 training sentences and 232 validation sentences.

Uspanteko was chosen for this research because it is a low-resource language that has little available data, but comprehensive linguistic analysis, and because it has a highly productive morphology.

We created a taxonomic tree of morphemes based on the gloss labels and \citet{coon_mayan_2016}, with the goal of creating classes where the glosses in a class tend to occur in similar contexts. The taxonomy is provided in \autoref{sec:taxonomy}.

\subsection{Inductive Biases}
This work falls into the category of techniques for encoding \textit{inductive biases}, aspects of the learning algorithm that create bias toward the desired type of solution \citep{battaglia_relational_2018}. In this case, we aim to create the bias that morphemes should be glossed with regards to a known taxonomy, and similar morphemes labels should be predicted similarly. 

\subsection{Taxonomic Loss Function}
The standard loss function for a token classification task is cross-entropy loss over the predicted output and a one-hot encoding of the correct label. When an incorrect label is predicted, this loss function penalizes the prediction equally regardless of if the predicted gloss is very close to the right answer or completely wrong. 

The taxonomic loss function modifies this by rewarding predicted glosses that have a close shared parent with the correct gloss on the taxonomic tree. This approach has been used in plant identification \citep{wu_deep_2019} and seafloor imagery classification \citep{nourani-vatani_structured_2015}, two other scenarios where there exists natural taxonomies of labels. The taxonomic loss function is calculated by summing cross-entropy loss at each level of classification, grouping logits for labels of the same class. The mathematical formulation is described in \autoref{sec:taxloss}.

\subsection{Model Architecture}
We first pre-trained a masked language model on the entire Uspanteko training corpus, since there is no existing language model to our knowledge. The model uses the RoBERTa architecture \citep{liu_roberta_2019}, but with three hidden layers, each of 100 dimension, and five attention heads. When the dataset is very small, \citet{gessler_microbert_2023} indicates that a much smaller language model architecture leads to better performance.

For token classification we replace the masked language modeling head of the pre-trained model with a single linear layer, and finetune on a subset of the training data. 

\section{Experimental Conditions}
Hyperparameters are provided in \autoref{sec:training_appendix}.

\begin{table*}[!htb]
    \centering
    \def\arraystretch{1.5}
    \begin{tabular}{l | c c c}
        \toprule
        & \textbf{Cross-entropy} & \textbf{Taxonomic} & \textbf{Harmonic taxonomic}\\
        \midrule
        10 & \textbf{26.3} & 24.1 & 24.8 \\
        100 & 30.0 & 32.9 & \textbf{34.8} \\
        500 & \textbf{47.3} & 45.7 & 43.1 \\
        1000 & \textbf{68.1} & 67.7 & 65.3 \\
        full & \textbf{86.5} & 86.0 & 85.9 \\
        \bottomrule
    \end{tabular}
    \caption{Average accuracy, for five different-sized datasets and three different loss functions with ten trials each}
    \label{tab:results}
\end{table*}

\begin{table*}[!htb]
    \centering
    \def\arraystretch{1.5}
    \begin{tabular}{l | c c c}
        \toprule
        & \textbf{Cross-entropy} & \textbf{Taxonomic} & \textbf{Harmonic taxonomic}\\
        \midrule
        10 & \textbf{55.3} & 51.3 & 50.4 \\
        100 & 64.2 & 62.8 & \textbf{67.5} \\
        500 & 80.4 & \textbf{81.7} & 81.5 \\
        1000 & 91.0 & 90.4 & \textbf{91.5} \\
        full & \textbf{97.9} & 97.8 & \textbf{97.9} \\
        \bottomrule
    \end{tabular}
    \caption{Average top-5 accuracy, as in \autoref{tab:results}}.
    \label{tab:top5results}
\end{table*}

\subsection{Fine-Tuning}
We fine-tuned models using three different loss functions: standard cross-entropy loss, taxonomic loss, and harmonic taxonomic loss. We trained on subsets of 10, 100, 500, and 1,000 labeled sentences to replicate the scenario where there is very limited labeled training data.

Since performance is very sensitive to the particular subset of training data, we trained each model ten times using different random seeds and averaged performance.

Fine-tuning used the same hyperparameters listed in \autoref{tab:pretrainhyper}, except we only trained for 100 epochs. Training took a total of roughly twenty hours.

\subsection{Evaluation}
Models were evaluated using the evaluation data set. For each model, we calculated the overall accuracy for every morpheme (ignoring word separators).

As one of the goals of this research is making predictions that are helpful to annotators, even if they are not exactly correct, we also calculated the \textit{top-5} accuracy, where predicting the correct gloss as one of the top five most likely glosses was counted as correct. In a human-in-the-loop setting, the automated glossing system might recommend a few likely glosses to the human annotator; this metric indicates how useful our system would be in that scenario.

\section{Results}

The average validation accuracy of ten trials is presented in \autoref{tab:results}. In all training sizes except 100, the standard cross-entropy loss model performs the best.

The average top-5 validation accuracy of ten trials is presented in \autoref{tab:top5results}. Here, the cross-entropy loss model only performs best for 10 training sentences, with the harmonic taxonomic loss model performing best for 100 and 1000 sentences, and the taxonomic loss model performing best for 500 sentences.

\section{Discussion}
\subsection{Taxonomic Loss}
In general, the taxonomic loss models did not show significant performance gains in overall accuracy. 

One key reason for this is while the taxonomic loss encourages shared representations for similar morpheme labels, it can worsen performance when distinguishing between two morphemes that are close taxonomically. A possible fix for this would be to gradually decrease the weighting for the loss of higher levels of taxonomy while training.

As the datasets are very small, the cross-entropy loss model may be relying more on heuristics that don't reflect actual morphological patterns but achieve better accuracy. To evaluate this, we could evaluate the model's ability to generalize to a dataset with a different distribution, such as a separate IGT corpus \citep{linzen_how_2020}.

That said, in individual runs at various dataset sizes, there were cases where the taxonomic or harmonic taxonomic loss models far outperformed the cross-entropy loss model. This demonstrates how performance is highly sensitive to the distribution of the training data when the dataset is very small. 

Furthermore, we see that the taxonomic and harmonic taxonomic loss models outperform the cross-entropy model for 100 training sentences. Considering this and the previous observation, a researcher building a system for glossing with very scarce data should consider using a taxonomic loss if a taxonomy is available.

\subsection{Taxonomy Construction}
One other potential major source of error is a sub-optimally constructed taxonomy. While we created the taxonomy here to the best of our ability, it's entirely possible that it misrepresents some finer details of the morphology.

The system is highly sensitive to the quality of the taxonomy. If an incorrect taxonomy is provided, the model will have difficulty learning to predict two glosses with a shared representation when those glosses do not actually share a similar context. Thus, future work should rely on a language expert to help construct a high-quality taxonomy.

\subsection{Taxonomic and Harmonic Taxonomic Loss}
We observe small differences in the performance of the taxonomic and harmonic taxonomic loss models, with the taxonomic model achieving better accuracy with larger training sets and vice versa. This result is consistent with the differences between the loss functions: the harmonic taxonomic loss weights smaller categorizations more heavily, allowing it to perform better on the glosses it observes in the training set, while the taxonomic loss induces deeper shared representations.

\subsection{Top-5 Accuracy}
The results for top-5 accuracy are very promising for the usage of taxonomic losses in a human-in-the-loop setting. One of the two taxonomic losses outperforms the standard cross-entropy loss at all training sizes except 10, by up to 3.3\%. While predicting the correct gloss with the top result is important for completely automated systems, in a system where the model recommends glosses to a human annotator, it is more important that the correct gloss is one of the top suggestions.

Furthermore, a qualitative analysis reveals that the taxonomic loss models make more useful top-5 predictions. For example, one particular token in the validation set should be glossed as the ergative \textsc{E1S}. The cross-entropy loss model predicts:

$$[\textsc{E1S}, \textsc{A1S}, \textsc{PRON}, \textsc{INC}, \textsc{[SEP]}]$$

While the taxonomic loss model predicts:
$$[\textsc{E3S}, \textsc{E1S}, \textsc{E1P}, \textsc{A2S}, \textsc{E2S}]$$

We see that while the taxonomic loss model gets the top prediction wrong, the top predictions it makes are closely related morphemes, which is both useful to the annotator and can inspire more trust in the system.

\section{Future Research}
While this research does not prove unequivocally that the taxonomic loss approach is superior, it does demonstrate that it can be effective in certain conditions, and it is worth researching it further.

Integrating morphological information into models through the loss function is highly sensitive to the quality of the taxonomic information. A more robust approach for integrating this same information might be using the taxonomy as input; for instance, a model trained to predict missing glosses based on context could encode the taxonomic information for the context glosses as input features.

\section{Conclusion}
In this paper, we demonstrated the novel use of the taxonomic loss function on language data, in the context of guiding morphological representation. We found that while these models do not necessarily outperform the standard amounts of data, they tend to suggest the top five glosses more accurately. This result suggests that taxonomic loss functions could be useful in a human-in-the-loop documentation setting, where the automated system provides multiple suggestions for a human annotator to choose from. Since linguistic analyses already exist for many languages, this is a promising approach to improve performance when data is very scarce. We hope that this result and future related research can aid in documentation and development of low-resource language systems, and ultimately can be one tool in the fight against language extinction.

\section{Limitations}
This research was conducted testing on a single language and corpus, and the effectiveness of the taxonomic loss function approach may vary for different languages and sources. In particular, for analytic languages and languages with minimal productive morphology (such as English), this technique will likely provide little benefit. 

The experiments utilized a single model architecture for consistency, but other architectures might show different performance for the various loss functions. We used a small transformer architecture due to the size of the training dataset, but a deeper network might show less benefit of using one loss function over another.

\section{Ethical Considerations}
While we feel the efforts to improve models for low-resource languages to be a net good, it is always worth considering the ethical implications. When working with low-resource languages, we must take care to avoid a colonialist approach that supercedes the will of the language community \citep{bird_decolonising_2020}.

In some cases, the language speaker community might not wish to preserve their native language, in favor of a more dominant one \citep{msila_mama_2014}. If this is the case, then documenting or making preservation efforts against the will of the community is inadvisable.

Additionally, relying on automated systems for documentation runs the risk of incorrect or insufficient documentation that ultimately loses information about the language, harming preservation efforts. Documentation should always be done in conjunction with experts and speakers of the language to ensure the meaning and cultural value of the texts are not diminished.

When using texts from a community, we should respect privacy and safety and remember that language is not just data in a vacuum. Corpora should always be collected and used in accordance with the speakers.

Although the models trained in this project were relatively small, training transformers using GPUs incurs an unavoidable carbon cost \citep{bender_dangers_2021}. We must take care to avoid unnecessary training of large models, and to mitigate their effects on the climate when possible.

\bibliography{anthology,custom}
\bibliographystyle{acl_natbib}

\appendix

\section{Experimental Details}
\label{sec:training_appendix}

\subsection{Preprocessing}
We encode each input morpheme as a unique integer, and add separator tokens between words. The data is otherwise sufficiently pre-processed for training.

\subsection{Training}
We pre-trained the masked language model with the parameters in \autoref{tab:pretrainhyper}. Training took roughly half an hour.

\begin{table}[!h]
    \centering
    \def\arraystretch{1.5}
    \begin{tabular}{l c}
        \toprule
        Parameter & Value \\
        \midrule
        Optimizer & AdamW  \\
        $\beta_1$ & 0.9 \\
        $\beta_2$ & 0.999 \\
        $\epsilon$ & $1\text{E}{-8}$ \\
        Weight decay & 0.01 \\
        Batch size & 64 \\
        Gradient accumulation steps & 3 \\
        Epochs & 200 \\
        GPU & NVIDIA V100 \\
        \bottomrule
    \end{tabular}
    \caption{Hyperparameters \\\small{\citep{adamw}}}
    \label{tab:pretrainhyper}
\end{table}

\section{Morphological Taxonomy}
\label{sec:taxonomy}
\begin{figure*}
{\tiny
\begin{forest} for tree={l sep=30pt,
                        inner sep=2pt,
                        grow=0,reversed, 
                        parent anchor=east,child anchor=west,}
    [Part-of-Speech 
        [Verb 
            [Abs/Erg 
                [Abs [Pl [A1P] [A2P]][Si [A1S] [A2S]]]
                [Erg [S/Pl [E1] [E2] [E3]][Pl [E1P] [E2P] [E3P]][Si [E1S] [E2S] [E3S]]]]
            [Transitiv. [AFE] [ITR] [TRN]]
            [Voice [AP] [APLI] [CAU] [MOV] [PAS] [REC] [RFX]]
            [TAM [Aspect [COM] [INC] [PRG]]
            [Mood [COND] [IMP] [INT]] [TAM]]
            [Deriv [DIR] [PP] [SC] [SV]]
            [Stem [EXS] [POS] [VI] [VT]]
        ]
        [Subs 
            [Stem [ADJ][NOM][PRON][S][SAB][TOP][VOC]]
            [Modifier [CLAS][DIM][GNT][NUM]]
            [Case [AGT][INS]]
            [Pl]
        ] 
        [Adv] [Partic [AFI][NEG][PART][SREL]] [Determ [ART][DEM]] [Conj] [Deriv [ENF][ITS][MED]] [Prep] ]
\end{forest}
}
\caption{Complete morphology taxonomy}
\label{fig:taxonomy}
\end{figure*}
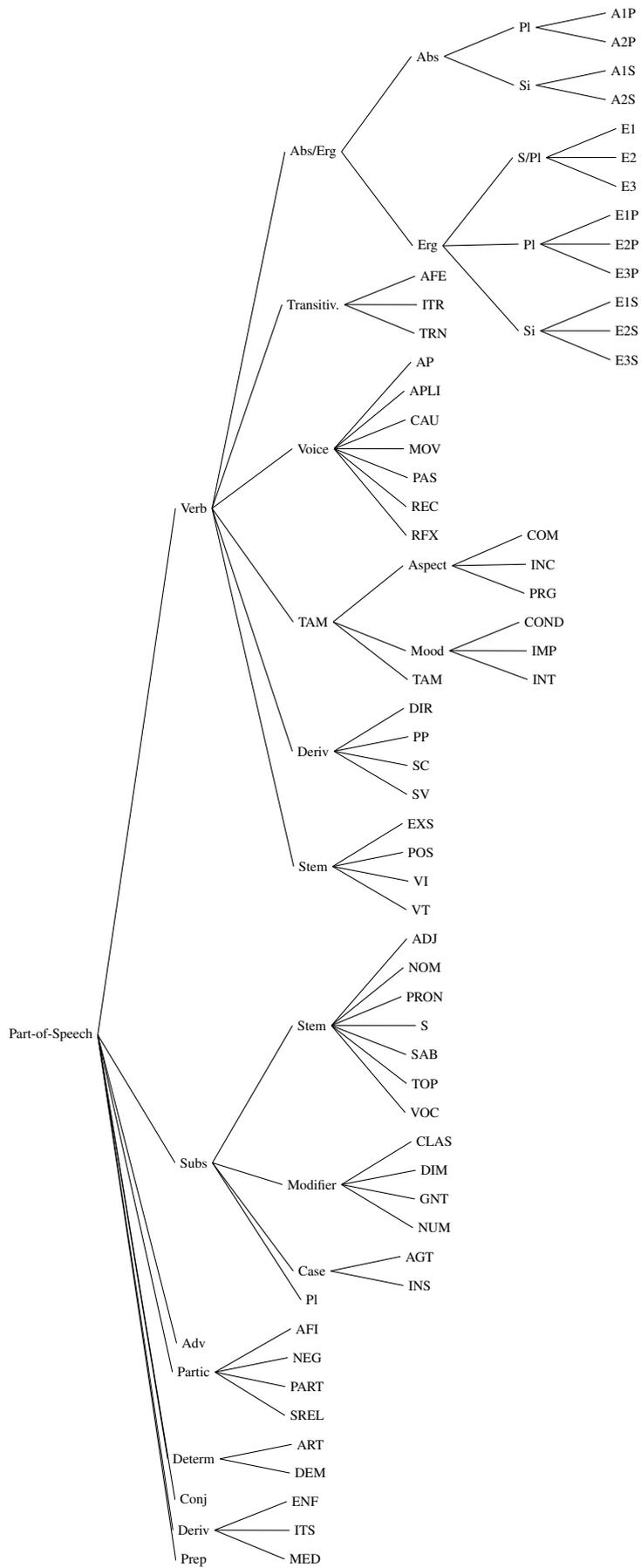

The taxonomy used is shown in \autoref{fig:taxonomy}.

Uspanteko uses a highly structured and productive morphology. For instance, we observe that Uspanteko tends to use the following order for verbal morphology:
$$\textsc{Tam}-\textsc{Abs}-\textsc{Erg}-\textsc{Root}-\textsc{Voice}-\textsc{Status}$$

Where \textsc{Tam} is tense-aspect-mood, \textsc{Abs} is absolutive, and \textsc{Erg} is ergative. We use these classes as part of our taxonomy and hope to capture this sort of pattern.

Our taxonomy encodes up to 5 levels of hierarchy and has 65 distinct glosses. For example, the gloss \textsc{E1P}, the ergative first-person plural, is categorized as $\textsc{V} \rightarrow \textsc{Abs/Erg} \rightarrow \textsc{Erg} \rightarrow \textsc{Pl} \rightarrow \textsc{E1P}$. In this way, \textsc{E1P} is taxonomically close to a gloss such as \textsc{E2P}, but is distant from a gloss like \textsc{Prep}.

\newpage

\begin{strip}
\begin{align}
    \label{taxloss}
\textsc{TaxLoss}(V, v^*, \hat{l}) = \sum_{j=1}^{j \leq d} \textsc{CELoss}\left(e_{v^*_j}, \left(\sum_{i = 1}^{i \leq n} \hat{l}_j [(V_i)_j = v^*_j]\right)e_{v^*_j} \right)
\end{align}
\begin{align}
\label{harmonictaxloss}
\textsc{HarmonicTaxLoss}(V, v^*, \hat{l}) = \sum_{j=1}^{j \leq d} \frac{\textsc{CELoss}\left(e_{v^*_j}, \left(\sum_{i = 1}^{i \leq n} \hat{l}_j [(V_i)_j = v^*_j]\right)e_{v^*_j} \right)}{d - j + 1}
\end{align}
\end{strip}

\section{Taxonomic Loss Function}
\label{sec:taxloss}

We calculate the taxonomic loss function as follows. Given some taxonomy of morpheme glosses with max depth $d$, create a vector $v \in \mathbb{R}^d$ for each gloss. The vectors are filled uniquely with ascending integers such that iff two glosses $v$ and $u$ are in the same class at level $i$ of the taxonomy, then $v_i = u_i$.

Let $V$ be the set of all morpheme gloss vectors, where $|V| = n$. For a single morpheme in the training sequence, let the correct gloss be $v^* \in V$ where $V_{i^*} = v^*$ for some $i^*$. Let the predicted logits be $\hat{l} \in \mathbb{R}^n$ where $\hat{l}_i = P(v = V_i)$ is the probability of some particular gloss label.

Traditional cross-entropy loss is calculated between the correct label and the predicted logits, which we denote as:
$$\textsc{CELoss}(e_{i^*}, \hat{l})$$

Taxonomic loss extends this by calculating the cross-entropy loss at each level of taxonomy and summing these losses (\ref{taxloss}). At each level $j$, we sum the logits for all the glosses in the same class of morphemes and then calculate the cross-entropy loss over these groups. The result of this function is that the more super-classes the predicted and correct glosses share, the lower the loss will be. Predicting a gloss which is a distant relative has much higher loss than predicting a gloss that is a close sibling. 

We also test a slight variation of this loss, where the loss from the final level of classification was weighted normally, the loss from the previous level of classification was divided by 2, the loss from the previous level was divided by 3, and so on. We call this \textit{harmonic taxonomic loss} (\ref{harmonictaxloss}).

\end{document}